\documentclass[10pt,twocolumn,letterpaper]{article}

\usepackage{iccv}
\usepackage{times}
\usepackage{epsfig}
\usepackage{amsmath}
\usepackage{amssymb}
\usepackage{pifont}
\usepackage{enumitem}
\usepackage{graphicx,subcaption}
\usepackage{amssymb,balance}
\usepackage{relsize}
\usepackage{url}
\usepackage{tabularx, booktabs}

\usepackage[table]{xcolor}
\usepackage{acronym}
\definecolor{lightgray}{gray}{0.9}
\definecolor{lightblue}{rgb}{0.93,0.95,1.0}
\definecolor{darkgreen}{rgb}{0.0,0.6,0.0}
\definecolor{mypink1}{rgb}{0.858, 0.188, 0.478}

\newcommand{\comment}[1]{}

\newcommand{\minisubsection}[1]{\vspace{1mm}{\textbf{#1}}}

\acrodef{NMI}{normalized mutual information}
\acrodef{ARI}{adjusted Rand index}
\acrodef{ACC}{clustering accuracy}
\acrodef{elu}{exponential linear unit}
\acrodef{KM}{K-means}
\acrodef{SC}{Spectral Clustering}
\acrodef{DAMIC}{Deep Autoencoder MIxture Clustering}

\newcommand{\ignore}[1]{}
\newcolumntype{L}[1]{>{\raggedright\let\newline\\\arraybackslash\hspace{0pt}}m{#1}}
\newcolumntype{C}[1]{>{\centering\let\newline\\\arraybackslash\hspace{0pt}}m{#1}}
\newcolumntype{R}[1]{>{\raggedleft\let\newline\\\arraybackslash\hspace{0pt}}m{#1}}
\DeclareTextFontCommand{\empha}{\bfseries\em}

\def\beq{\begin{equation}}
\def\eeq{\end{equation}}

\def\beqary{\begin{eqnarray}}
\def\eeqary{\end{eqnarray}}

\def\beqarz{\begin{eqnarray*}}
\def\eeqarz{\end{eqnarray*}}

\iccvfinalcopy  

\def\iccvPaperID{3803} 

\ificcvfinal\fi
\begin{document}

\title{ Deep Clustering based on a Mixture of Autoencoders}

\author{Shlomo E. Chazan, Sharon Gannot and Jacob Goldberger\\
Bar-Ilan University\\
 Ramal-Gan, 5290002, Israel\\
{\tt\small \{Shlomi.Chazan, Sharon.Gannot, Jacob.Goldberger\}@biu.ac.il}}

\maketitle

\begin{abstract}

In this paper we propose a Deep Autoencoder Mixture Clustering (DAMIC) algorithm based on a mixture of deep autoencoders where each cluster is represented by an autoencoder. A clustering network transforms the data into another space and then selects one of the clusters. Next, the  autoencoder associated with this cluster is used to reconstruct the data-point. The clustering algorithm jointly learns the  nonlinear data representation and the set of autoencoders.
The optimal clustering is found by minimizing the reconstruction loss of the mixture of autoencoder network. Unlike other deep clustering algorithms, no regularization term is needed to avoid data collapsing to a single point.
Our experimental evaluations on image and text corpora show significant improvement over state-of-the-art methods.
\end{abstract}

\section{Introduction}
Effective automatic grouping of objects into clusters is one of the fundamental
problems in machine learning and data analysis. In many approaches,
the first step toward clustering a dataset is extracting a feature vector from each
object. This reduces the problem to the aggregation of groups of vectors in a
feature space. A commonly used clustering algorithm in this case is  $k$-means.
Clustering high-dimensional datasets is, however, hard  since the inter-point distances become
less informative in high-dimensional spaces. As a result, representation learning has been
 used  to map the input data into a low-dimensional feature space.
In recent years, motivated by the success of deep neural networks in supervised learning,
there have been many attempts to apply unsupervised deep learning approaches to clustering.
Most methods  are focused on clustering over the low-dimensional feature space of an autoencoder  \cite{yang2016joint}\cite{dizaji2017deep}\cite{hu2017learning}\cite{yang_2017}, a variational autoencoder \cite{vade} \cite{gmvae} or a Generative adversarial Network (GAN) \cite{gan1}\cite{gan2}\cite{gan3}.
Recent good overviews of  deep clustering  methods can be found in \cite{ Aljalbout} and  \cite{Erxue}.

Using deep neural networks,  nonlinear mappings that can transform the data into more clustering-friendly
representations, can be learned.
A deep version of $k$-means is based on learning a nonlinear data representation
and applying $k$-means in the embedded space. A straightforward implementation of the deep $k$-means algorithm would lead, however,
 to a trivial solution where  the features are collapsed to a single point in the embedded space  and the centroids are collapsed into a single entity. For this reason, the objective function of most deep clustering algorithms is composed of a clustering term computed in the embedded space and a regularization term in the form of a
 reconstruction error to avoid data collapse.
Deep Embedded Clustering (DEC) \cite{Xie_2015}  is first pre-trained  using an autoencoder reconstruction
loss and then  optimizes cluster centroids in the embedded  space through  a Kullback-Leibeler divergence loss. The Deep Clustering Network (DCN)  \cite{yang_2017} is another autoencoder-based method that uses $k$-means for clustering.
Similar to DEC, in the first phase, the network is pre-trained using the autoencoder reconstruction
loss. However, in the second phase, in contrast to DEC, the network is jointly trained using
a mathematical combination of the autoencoder reconstruction loss and the $k$-means clustering loss
function. Thus, because strict cluster assignments were used during the training (instead
of probabilities such as in DEC) the method requires an alternation process between  network
training and  cluster updates.

In this paper we propose an algorithm to perform deep  clustering within the mixture-of-experts framework \cite{mixture_of_experts}.
Each cluster is represented by an  autoencoder neural-network and the clustering itself is performed in a low-dimensional embedded space
 by a softmax classification layer that directs  the input data to the most suitable autoencoder.  Unlike most deep clustering algorithms the proposed algorithm is deep in nature and not a deep variant of a classical clustering algorithm.
The proposed deep clustering approach is  different from previous algorithms in three main aspects:
\begin{itemize}
\item  It does not suffer from the clustering collapsing problem, since the trivial solution is not the global optimum of the clustering learning objective function.
\item This implies that in the proposed method, unlike other methods,  there is no need for regularization terms that have to be tuned separately for each dataset. Note that parameter
tuning in clustering is problematic since it is based, either explicitly or implicitly, on the data labels
which are supposedly unavailable in the clustering process.
\item Another major difference between the proposed method and previously proposed approaches is the learning method of the embedded latent space, where the actual clustering takes place. In most previous methods, the embedded space is controlled by an autoencoder. Thus, in order to gain a good reconstruction, it requires to encode into the embedded space information that can be entirely irrelevant to the clustering process. In contrast, in our algorithms no decoding is applied to the clustering in the embedded space and the only goal of the embedded space is to find a good organization of the data into separated clusters.
\end{itemize}

We validate the method using standard real datasets including  document and image corpora. The results show a  visible improvement from previous methods  for all the datasets.
The contribution of this paper is thus twofold: (i) it presents a novel deep learning clustering method that unlike  deep variants of $k$-means does not require a tuned regularization term to avoid clustering collapse to a single point; and (ii)
it demonstrates improved performance on standard datasets.

\section{Mixture of Autoencoders}

Consider the problem of clustering a set of $n$ points $x_1,\ldots,x_n\in R^d$ into $k$ clusters.
The $k$-means algorithm represents each cluster by a centroid.
In our approach, rather than representing a cluster by a centroid, we represent each cluster by an autoencoder that is specialized in
reconstructing objects belonging to that cluster. The clustering itself is carried out by directing the input object to the most suitable autoencoder.

We next formally describe the proposed clustering algorithm.
 The algorithm is based on a (soft) clustering network  that produces a distribution over  the $k$ clusters:
\begin{equation}
p(c=i|x;\theta_c) =  \frac{\exp(w_i h(x)+b_i) }{ \sum_{j=1}^k \exp(w_j h(x)+b_j) }, \hspace{0.3cm} i=1,\ldots,k
\label{cluster_net}
\end{equation}
such that $\theta_c$ is the parameter set of the clustering network, $h(x)$ is a nonlinear representation of a point $x$  computed by the clustering network and
$w_1,\ldots, w_k,b_1,\ldots,b_k \in \theta_c$ are the parameters of the softmax output layer.
 The (hard) cluster assignment of a point $x$ is thus:
\begin{equation}
\hat{c} = \arg \max_{i=1}^k  p(c=i|x;\theta_c)  =  \arg \max_{i=1}^k  (w_i h(x)+b_i).
\label{clustering_assign}
\end{equation}
The clustering task is, by definition, unsupervised and therefore we cannot directly train the clustering network.
Instead, we use the clustering results to obtain a more  accurate reconstruction of the network input.
 We represent each cluster by an autoencoder that is specialized in reconstructing instances of that cluster.  If the dataset is properly clustered, we expect  all the points  assigned to be same cluster to be similar. Hence, the task of a cluster-specialized  autoencoder should be relatively easy compared to using a single autoencoder for the entire data.
    We thus expect that  good clustering should result in a small reconstruction error.
  Denote the autoencoder associated with  cluster $i$ by $f_i(x;\theta_i)$  where $\theta_i$ is the parameter-set of the network autoencoder. We can view the reconstructed object $f_i(x;\theta_i)\in R^d$ as a data-driven centroid of  cluster $i$ that is tuned to the input $x$. The goal of the training procedure is to find a clustering of the data such that the error of the cluster-based reconstruction is minimized.

To find the network parameters we jointly train the clustering network and the deep autoencoders. The clustering is thus computed by minimizing the following loss function:
\begin{equation}
L(\theta_1,\ldots,\theta_k,\theta_c)
\label{clustering_loss}
\end{equation}
$$ = - \sum_{t=1}^n \log \left( \sum_{i=1}^k p(c_t=i|x_t;\theta_c)  \exp( -d(x_t, f_i(x_t;\theta_i))) \right)$$

such that $d(x_t, f_i(x_t;\theta_i))$  is the reconstruction error of the $i$-th autoencoder. In our implementation we set $d(x_t, f_i(x_t;\theta_i))=\frac{1}{2}\|x_t- f_i(x_t;\theta_i)\|^2$.

In the minimization of (\ref{clustering_loss}) we simultaneously perform data clustering in the embedded space $h(x)$ and learn a `centroid' representation for each cluster in the form of an autoencoder.
Unlike most  previously proposed deep clustering methods,  there is no risk of collapsing to a
trivial solution where all the data points are transformed to the same vector, even though the clustering is carried out in the embedded space.  Collapsing all the data points into a single vector in the embedded space will result in directing all the points to the same autoencoder for reconstruction. As our clustering goal is to minimize the reconstruction error, this situation is, of course, worse than using  $k$ different autoencoders for reconstruction.
Hence, there is no need to add regularization terms to the loss function (that might influence the clustering accuracy) to prevent data collapse. Specifically, there is no need to add a decoder to the embedded space where the clustering is actually performed to prevent data collapse.

The   back-propagation equation for the parameter set of the clustering  network  is:
\begin{equation}
\frac{\partial L}{\partial {\theta}_{c}} = -\sum_{t=1}^n  \sum_{i=1}^k {w}_{ti} \cdot \frac{\partial }{\partial \theta_c} \log  p(c_t=i|x_t;\theta_c)
\label{dmoefder2}
\end{equation}
such that
\begin{equation}
w_{ti}=  \frac { p(c_t=i|x_t;\theta_c)  \exp (-d(x_t, f_i(x_t;\theta_i)))}
{\sum_{j=1}^k p(c_t=j|x_t;\theta_c)  \exp (-d(x_t, f_j(x_t;\theta_j)))}
\label{sestep}
\end{equation}
is a soft assignment of $x_t$ into the $i$-th cluster based on the current parameter-set.
In other words, the reconstruction error of the autoencoders is used to obtain soft labels that are employed for   training the clustering network.
\begin{figure*}
    \centering
    \includegraphics[trim = 10 100 10 50, clip, scale=0.2]{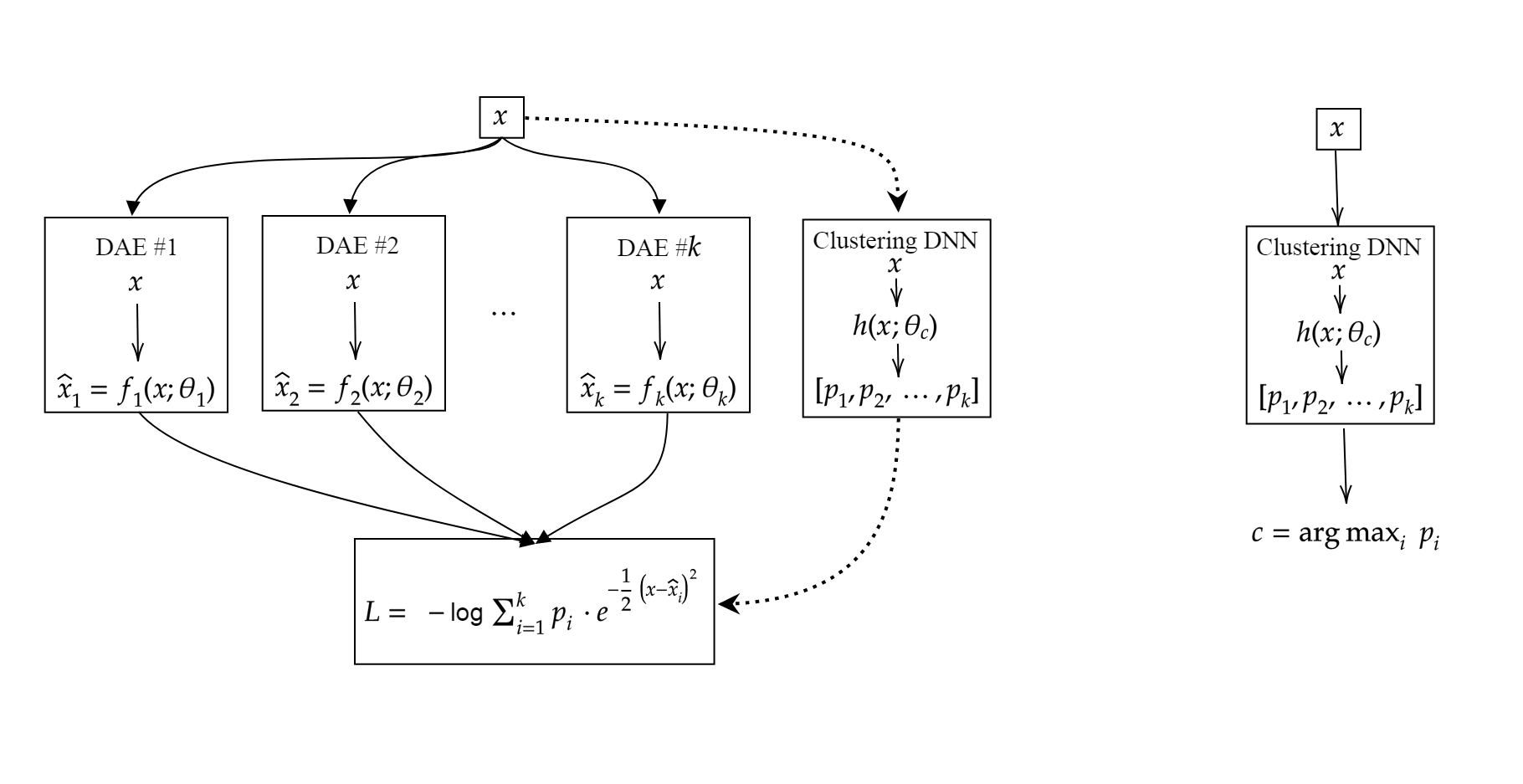}
    \caption{A block diagram of the proposed mixture of deep auto-encoders for clustering. The training procedure is described on the left side, and the final clustering on the right side.}
    \label{fig:block_diagram}
\end{figure*}

In recent years, network  pre-training has been largely rendered obsolete for supervised tasks, given availability of large labeled training datasets.
However, for hard optimization problems that unsupervised clustering tasks cannot handle (like the one presented in \eqref{cluster_net}), initialization is still crucial. To initialize
the parameters of the network, we first train a single autoencoder and  use the
layer-wise pre-training method, as described in \cite{Bengio_2007}, for training autoencoders.
After training the autoencoder, we carry out a $k$-means clustering on the output of the bottleneck layer to obtain the initial clustering values.
The $k$-means assigns a label to each data point. Note, that in the pre-training procedure a \emph{single} autoencoder is trained on the entire database. We use these labels as  supervision to pre-train the clustering network (\ref{cluster_net}).
The points that were assigned by the $k$-means algorithm to cluster $i$, are next used to pre-train the $i$-th autoencoder $f_i(x;\theta_i)$.
Once all the network parameters have been initialized by this pre-training procedure, the network parameters are jointly trained to minimize the
autoencoding reconstruction error defined by the loss function (\ref{clustering_loss}).
We dub the proposed algorithm  \acf{DAMIC}.  The architecture of the network  trained by the \ac{DAMIC} algorithm and the final clustering procedure are depicted in the left and right panels of Fig.~\ref{fig:block_diagram}, respectively. The clustering algorithm is summarized in Table \ref{table:algorithm}.

The \ac{DAMIC} algorithm can be viewed as an extension of the $k$-means algorithm. Assume  we replace each autoencoder in our network by  a constant function $f_i(x_t,\theta_i) \equiv \mu_i \in R^d$ and we replace the clustering network by a hard decision based on the reconstruction error. In so doing, we obtain exactly the classical $k$-means algorithm. The \ac{DAMIC} algorithm replaces the constant centroid with a data driven representation of the input computed by an autoencoder.

The probabilistic modeling used by the \ac{DAMIC} clustering algorithm can also be viewed as an instance of  the mixture-of-experts (MoE) model
 introduced in \cite{mixture_of_experts} and \cite{jordan1994hierarchical}.
 The MoE model is comprised of several
expert models and a gate model. Each of the experts provides a decision and the
gate is a latent variable that selects the relevant expert based on the input data.
In spite of the huge success of deep learning, there are only a few studies that have explicitly utilized
and analyzed MoEs as an architectural component of a neural network \cite{Eigen,Shazeer}.
MoE has been primarily applied to supervised tasks such as classification and regression.
In our clustering algorithm the clustering network is the equivalent of the MoE gating function.
The experts here are autoencoders were  each autoencoder's expertise is t reconstruct a sample from the associated  cluster.     Our  clustering cost function (\ref{clustering_loss}) follows the training strategy proposed in \cite{mixture_of_experts}, which  prefers an error function that encourages expert specialization instead of cooperation.

\begin{table}[h]
\small
\caption{The \acf{DAMIC} algorithm.}
\centerline{\fbox{\parbox{0.95\linewidth}{ \vskip0.2cm
Goal: clustering $x_1,\ldots,x_n\in R^d$ into $k$ clusters. \\ \\
Network components:
\begin{itemize}
\item 
A network that computes a soft clustering of   the data point: $$p(c=i|x;\theta_c) =  \frac{\exp(w_i h(x)+b_i) }{ \sum_{j=1}^k \exp(w_j h(x)+b_j) }$$
\item A set of autoencoders (one for each cluster): $$x \rightarrow \hat{x}_i =f_i(x;\theta_i), \hspace{0.5cm} i=1,\ldots,k$$
\end{itemize}
\vspace{0.1cm}
Pre-training:
\begin{itemize}
    \item Train a single autoencoder for the entire dataset.
    \item Apply a $k$-means algorithm in the embedded space.
    \item Use the $k$-means clustering to initialize the network parameters.
\end{itemize}
\vspace{0.5cm}
Training: clustering is obtained by minimizing the reconstruction error:
$$L(\theta_1,\ldots,\theta_k,\theta_c) = $$ $$ - \sum_{t=1}^n \log \big( \sum_{i=1}^k p(c_t=i|x_t;\theta_c)  \exp (-d(x_t, f_i(x_t;\theta_i))) \big)$$
\vspace{0.1cm}
The final (hard) clustering is:
$$\hat{c}_t = \arg \max_{i=1}^k  p(c_t=i|x_t;\theta_c), \hspace{0.5cm} t=1,\ldots,n.$$
}}}
\label{table:algorithm}
\end{table}

We note that after the training  process is finished, there is another way to extract the clustering from the trained network. Given a data point $x_t$, we can ignore the clustering DNN  and assign each point to the cluster whose reconstruction error is minimal:
\begin{equation}{c}_t = \arg \min_{i=1}^k  d(x_t, f_i(x_t;\theta_i))).
\end{equation}
We found that the performance of this clustering decision is very close to the clustering strategy we proposed (\ref{clustering_assign}). Moreover, in almost all cases the hard classification decision of the
clustering network (\ref{cluster_net})
coincides with the cluster whose reconstruction error is minimal, i.e., $$
\arg \max_{i=1}^k  p(c_t=i|x_t;\theta_c) = \arg \min_{i=1}^k  d(x_t, f_i(x_t;\theta_i))).
$$
We can thus consider a variant of our clustering algorithm that completely avoids the clustering network. Instead, the training goal is to directly minimize the reconstruction error of the most suitable autoencoder using the following cost function:
\begin{equation}
L(\theta_1,\ldots,\theta_k) =  \sum_{t=1}^n \min_{i=1}^k  d(x_t, f_i(x_t;\theta_i))).
\label{alt_cost}
\end{equation}
This cost function is very similar to the cost function of the $k$-means algorithm. The only difference is that the constant centroid is replaced here by the autoencoder bottleneck where the given point is the input.
However, there are two drawbacks of using this alternative  and simpler cost function. First, in our algorithm, in addition to the data clustering, we also obtain a nonlinear data embedding  $x \rightarrow h(x)$  that can be used to visualize the clustering in a clustering friendly space. The second issue is that we found empirically that  without the clustering network even if we use the pre-processing procedure we described above, we are more vulnerable to  clustering collapsing issues, in the sense that at the end of the training procedure some of  the autoencoders are not used by any data point.  This provides another motivation for the proposed architecture that is based on an explicit  clustering network.

\begin{figure*}[t]
    \centering
    \begin{subfigure}{0.35\textwidth}
    \centering \includegraphics[scale=0.35]{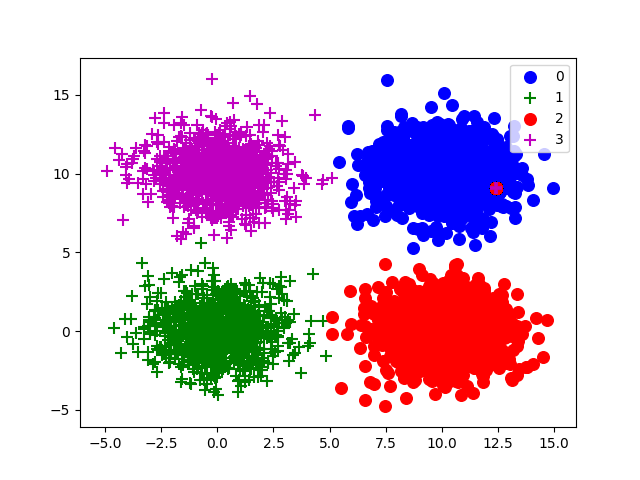}
    \caption{Latent domain, $v$.}
    \label{fig:synthetic_low_dim}
    \end{subfigure}%
    \begin{subfigure}{0.35\textwidth}
    \centering \includegraphics[scale=0.35]{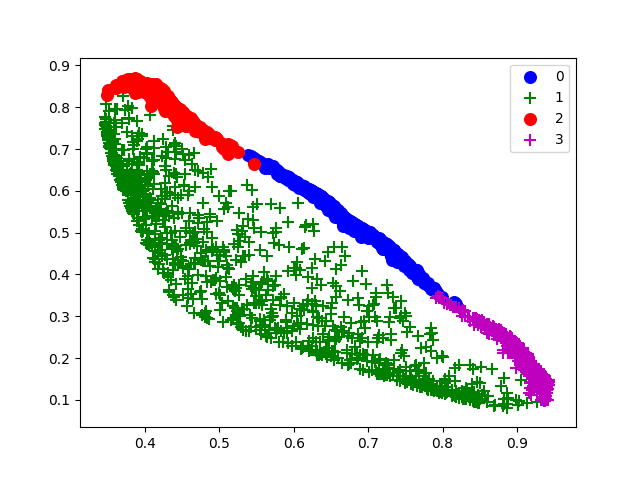}
    \caption{NMF}
    \label{fig:synthetic_nmf}
    \end{subfigure}%
    \begin{subfigure}{0.35\textwidth}
    \centering \includegraphics[scale=0.35]{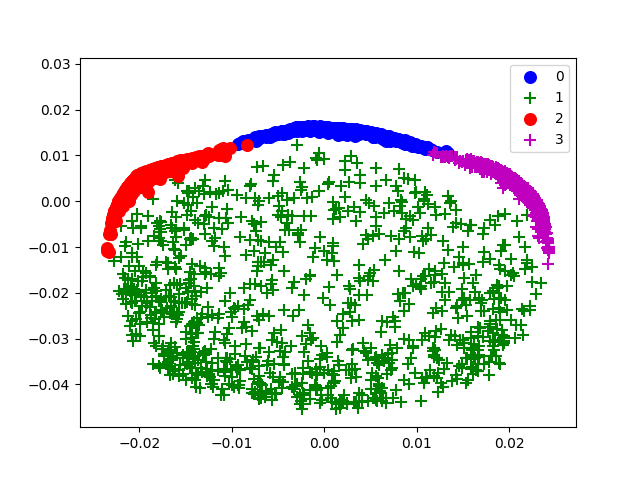}
    \caption{SVD}
    \label{fig:synthetic_svd}
    \end{subfigure}//

     \begin{subfigure}{0.35\textwidth}
    \centering \includegraphics[scale=0.35]{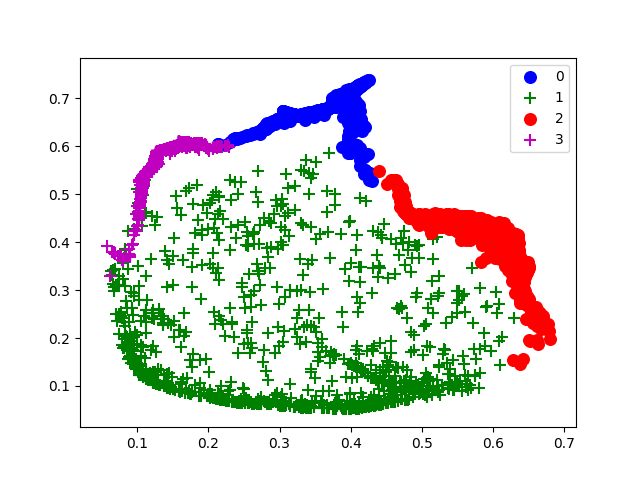}
    \caption{DAE+KM}
    \label{fig:synthetic_DAE_KM}
    \end{subfigure}%
      \begin{subfigure}{0.35\textwidth}
    \centering \includegraphics[scale=0.35]{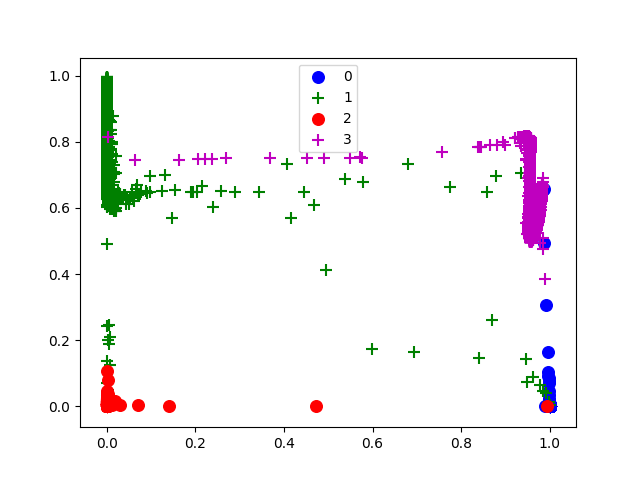}
    \caption{DAMIC}
    \label{fig:synthetic_DAMIC}
    \end{subfigure}
        \caption{Synthetic dataset with 4 clusters. Each true cluster label has a different color.  The observable data is generated from the Gaussian distributed clusters in the first figure, through  \eqref{synthetic_input}. The 2D representations of the observed data are shown by the NMF, SVD, DAE+KM and the proposed DAMIC methods.}
    \label{fig:synthetic}
\end{figure*}

\section{Experiments}

In this section we evaluate the clustering results of our approach. We carried out experiments on different
datasets and compared the proposed method to the state-of-the-art standard and $k$-means related deep clustering algorithms.

\subsection{Datasets} We used both synthetic dataset as well as real datasets. The synthetic dataset will be described in Sec.~\ref{sec:synthetic_dataset}.
The real datasets used in the experiments are standard clustering benchmark collections. We considered both image and text datasets to demonstrate the general applicability of our approach. 

The image datasets consisted of MNIST (70,000 images, 28 $\times$ 28 pixels, 10 classes)  which contain hand-written digit images. We reshaped the images to one dimensional
vectors and normalized the pixel intensity levels (between 0 and 1). The Fashion dataset~\cite{fashion}, which is consisting of 70,000 examples with similar dimensions as of the MNIST dataset, was also used. This dataset is divided into 10 fashion classes. 

The text collections we considered are the 20 Newsgroups dataset
(hereafter 20NEWS) and the RCV1-v2 dataset (hereafter  RCV1)~\cite{lewis2004rcv1}. For 20NEWS, the entire dataset comprising 18,846 documents labeled into 20 different news-groups was used. For the RCV1, similar to \cite{yang_2017} we used  a subset of the database containing  365,968 documents, each of which pertains to only one of 20 topics. Because of  text dataset sparsity, and as proposed in \cite{Xie_2015} and \cite{yang_2017}, we selected the 2000 words with the highest tf-idf values to represent each document.

\subsection{Evaluation measures} The clustering performance of the methods was evaluated with respect to the following three
standard measures: \acf{NMI} \cite{Cai}, \acf{ARI} \cite{Yeung},
and \acf{ACC} \cite{Cai}.
 NMI is an
information-theoretic measure based on the mutual information of the ground-truth classes and the obtained
clusters, normalized using the entropy of each.
ACC measures the proportion of data points for which the obtained clusters can be correctly
mapped to ground-truth classes, where the matching is based on the Hungarian algorithm \cite{ACC}.
Finally, ARI is a variant of the Rand index that is adjusted for the chance grouping of elements.
Note, that NMI and
ACC lie in the range of $0$ to $1$ where  one is the perfect
clustering result and zero the worst. ARI is a value
between minus one to (plus) one, where one  is the best clustering performance and minus one  the worst.

\subsection{ Baseline methods}
The proposed DAMIC algorithm was compared to the following methods:
\begin{description}
    \item[\acf{KM}:] The classic $k$-means \cite{Kmeans}.
    \item [\acf{SC}:] The classic SC algorithm \cite{spectral_clustering}.
    \item[Deep Autoencoder followed by $k$-means (DAE+\ac{KM}):] This algorithm is carried out in two steps. First, a DAE is applied. Next, \ac{KM} is applied to the embedded layer of the DAE. This algorithm is also used as an initialization step for the
 proposed algorithm.
 \item[Deep Clustering Network (DCN):] The algorithm performs joint reconstruction and $k$-means clustering at the same time. The loss comprises penalties on both the reconstruction and the clustering losses \cite{yang_2017}.
 \item[Deep Embedding Clustering (DEC):]  The algorithm performs joint embedding and clustering in the embedded space. The  loss function only contains  a clustering loss term \cite{Xie_2015}.
\end{description}

\subsection{Network implementation}
The proposed method was implemented with the deep learning toolbox Tensorflow \cite{abadi2016tensorflow}. All datasets were normalized between 0 and 1. All neurons in the proposed architecture except the output layer used  \ac{elu} as the transfer function. The output layer in all DAEs was the sigmoid function, and the clustering network output layer was a softmax layer. Batch normalization \cite{batchnorm}  was utilized on all layers, and the ADAM optimizer \cite{adam} was used for both the pre-training as well as the training phase. In the pre-training phase, the DAE networks were trained with the binary cross-entropy loss function. We set the number of epochs for the training phases to be 50. However, \emph{early stopping} was used to prevent mis-convergence of the loss. The mini-batch size was 256.

Note that for simplicity and to show the robustness of the proposed method, the architectures of the proposed \ac{DAMIC} in all the following experiments had a similar shape; i.e., for each of the DAEs we used a 5-layer DNN with the following input size: 1024, 256, $k$, 256, 1024, \ac{elu} neurons, respectively,   and for the clustering network we used 512, 512, $k$, \ac{elu} neurons, respectively, where $k$ is the number of clusters. There was no need for hyperparameter tuning  for the  experiments on the different datasets.

\section{Results}

\subsection{Synthetic dataset} \label{sec:synthetic_dataset}
To  illustrate the capabilities of the DAMIC algorithm we generated  synthetic data as in \cite{yang_2017}.  The 2D latent domain  contained 4000 samples from four Gaussian distributed clusters as shown in Fig.~\ref{fig:synthetic_low_dim}.  The observed signal is
\begin{equation}
    x_t=(\sigma(\mathbf{W}\cdot v_t))^2\quad t=1,\cdots,n
    \label{synthetic_input}
\end{equation}
where  $\sigma$ is the sigmoid function, $\mathbf{W}\in \mathbb{R}^{100\times2}$ and $v_t$ is the $t$-th point in the latent domain.

We first applied the DAE+KM algorithm for initialization. The architecture of the DAE consisted of a 4-layer encoder with 100, 50, 10, 2 neurons respectively. The decoder was a mirrored version of the forward network. Fig.~\ref{fig:synthetic_nmf}, \ref{fig:synthetic_svd} and \ref{fig:synthetic_DAE_KM} depict the 2D representations of \eqref{synthetic_input} by NMF, the SVD and the DAE+KM methods, respectively. It is clear that it is not sufficiently separated.

The proposed DAMIC algorithm was then applied. The architecture of each autoencoder consisted of 5-layers of 1024, 256, 4, 256, 1024 neurons as described in the previous section. The clustering network was also similar, with 512, 512, 2 neurons, respectively.   Fig.~\ref{fig:synthetic_DAMIC} depicts the 2D embedded space of the clustering network $h(x_t)$. It is easy to see that the embedded space is much more separable.

Table~\ref{table:synthetif_results} summarizes the results of the $k$-means,  the DAE+KM, the \ac{SC} and the DAMIC algorithms on the  synthetic generated data. It is easy to verify that the DAMIC algorithm outperforms the two competing algorithms in both NMI and ARI measures.

\begin{table}[h]
  \centering
    \caption{Objective measures for the synthetic database.}\label{table:synthetif_results}
\begin{tabular}{l|*{4}c}
\hline
Method            &  \ac{DAMIC} &  DAE+KM  & SC &KM  \\
\hline
NMI              & \textbf{0.94} & 0.83 & 0.82 & 0.80   \\
ARI              & \textbf{0.96} & 0.84 & 0.83 & 0.81   \\
\hline
\end{tabular}
\end{table}
\begin{table}
  \centering
   \caption{Objective measures of the MNIST database.}\label{table:mnist_results}
\begin{tabular}{l|*{6}c}
\hline
Method            &  \ac{DAMIC} & DCN & DAE+KM  & DEC & KM  \\
\hline
NMI              & \textbf{0.87} & 0.81 & 0.74 & 0.80 & 0.50    \\
ARI              & \textbf{0.81} & 0.75 & 0.67 & 0.75 &  0.37   \\
ACC              & \textbf{0.89} & 0.83 & 0.80 & 0.84 &  0.53   \\
\hline
\end{tabular}
 \end{table}
 \begin{table}
  \centering
   \caption{Objective measures of the Fashion database.}\label{table:fashion_results}
\begin{tabular}{l|*{6}c}
\hline
Method            &  \ac{DAMIC} & DCN & DAE+KM  & DEC & KM  \\
\hline
NMI              & \textbf{0.65} & 0.55 & 0.60 & 0.54 & 0.51    \\
ARI              & \textbf{0.49} & 0.42 & 0.45 & 0.40 &  0.37   \\
ACC              & \textbf{0.60} & 0.50 & 0.57 & 0.51 &  0.47   \\
\hline
\end{tabular}
 \end{table}
\begin{table}
  \caption{Objective measures of the 20NEWS database.}\label{table:20newsgroup_results}
  \centering
\begin{tabular}{l|*{6}c}
\hline
Method            &  \ac{DAMIC} & DCN & DAE+KM  & SC & KM & \\
\hline
NMI              & \textbf{0.57} & 0.48 & 0.47 & 0.40  & 0.41   \\
ARI              & \textbf{0.42} & 0.34 & 0.28 & 0.17 &  0.15   \\
ACC              &  \textbf{0.56} &  0.44 & 0.42  &  0.34 &0.30     \\
\hline
\end{tabular}
\end{table}
\begin{table}
  \centering
    \caption{Objective measures of the RCV1 database.}\label{table:RCV1_results}
\begin{tabular}{l|*{6}c}
\hline
Method            &  \ac{DAMIC} & DCN & DAE+KM &DEC  & KM & \\
\hline
NMI              & \textbf{0.62} & 0.61 & 0.59 &  0.08 & 0.58   \\
ARI              & \textbf{0.41} & 0.33 & 0.33 & 0.01 &  0.29   \\
ACC              &  \textbf{0.47} &  \textbf{0.47} & 0.46  & 0.14  &0.47     \\
\hline
\end{tabular}
\end{table}

\begin{figure*}[t]
    \centering
    \begin{subfigure}{0.2\textwidth}
    \centering \includegraphics[scale=0.15]{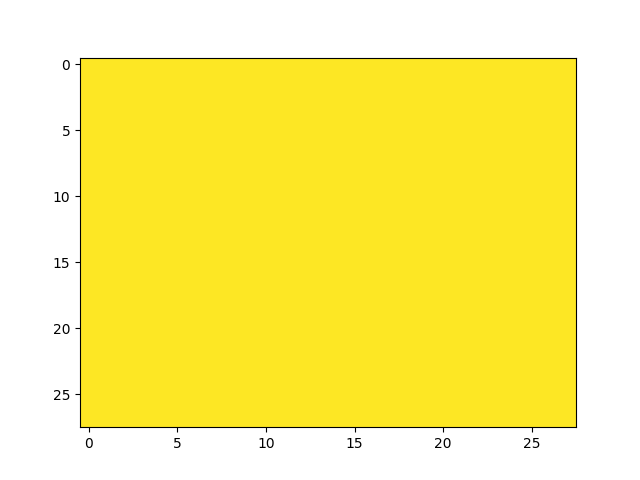}
    \caption{Input}
    \label{fig:all_one}
    \end{subfigure}%

     \begin{subfigure}{0.2\textwidth}
    \centering \includegraphics[scale=0.15]{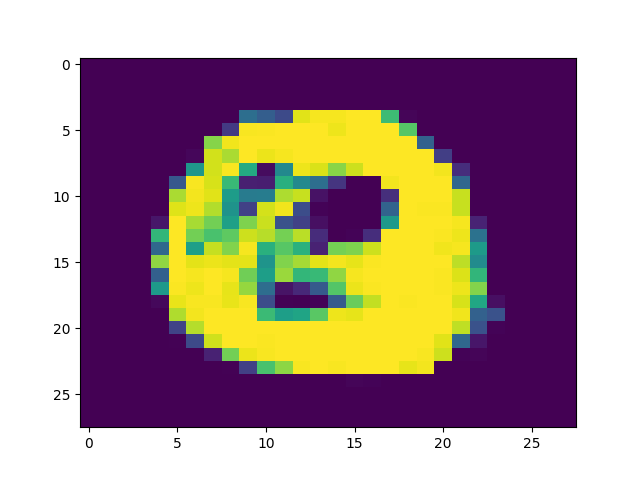}
    \caption{DAE \#0}
    \end{subfigure}%
     \begin{subfigure}{0.2\textwidth}
    \centering \includegraphics[scale=0.15]{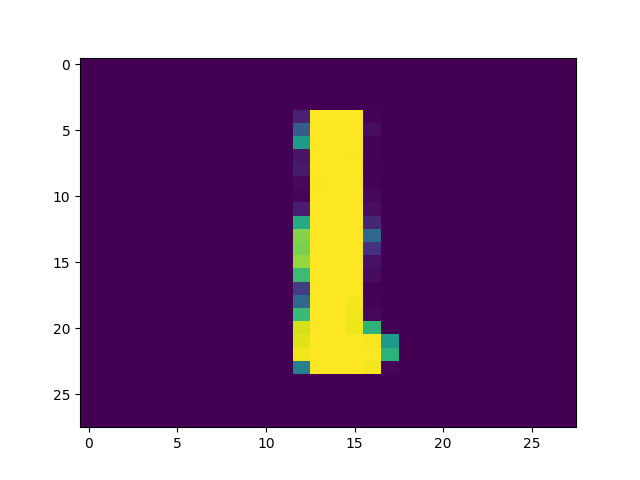}
    \caption{DAE \#1}
    \end{subfigure}%
     \begin{subfigure}{0.2\textwidth}
    \centering \includegraphics[scale=0.15]{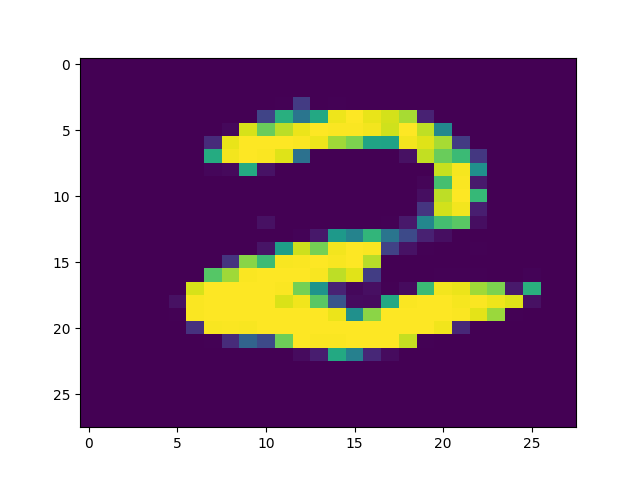}
    \caption{DAE \#2}
    \end{subfigure}%
    \begin{subfigure}{0.2\textwidth}
    \centering \includegraphics[scale=0.15]{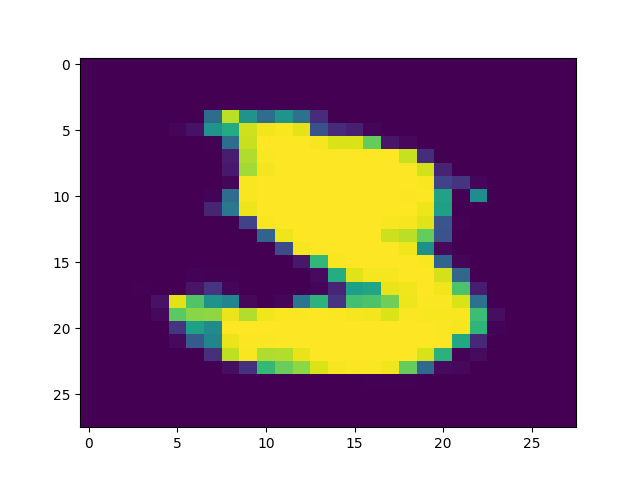}
    \caption{DAE \#3}
    \end{subfigure}%
     \begin{subfigure}{0.2\textwidth}
    \centering \includegraphics[scale=0.15]{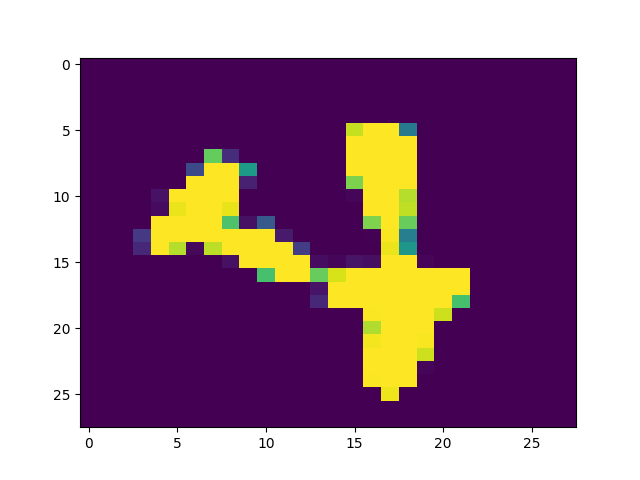}
    \caption{DAE \#4}
    \end{subfigure}%

    \begin{subfigure}{0.2\textwidth}
    \centering \includegraphics[scale=0.15]{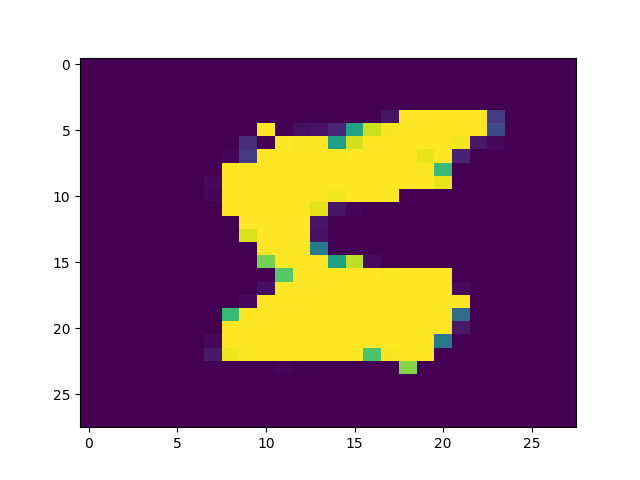}
    \caption{DAE \#5}
    \end{subfigure}%
     \begin{subfigure}{0.2\textwidth}
    \centering \includegraphics[scale=0.15]{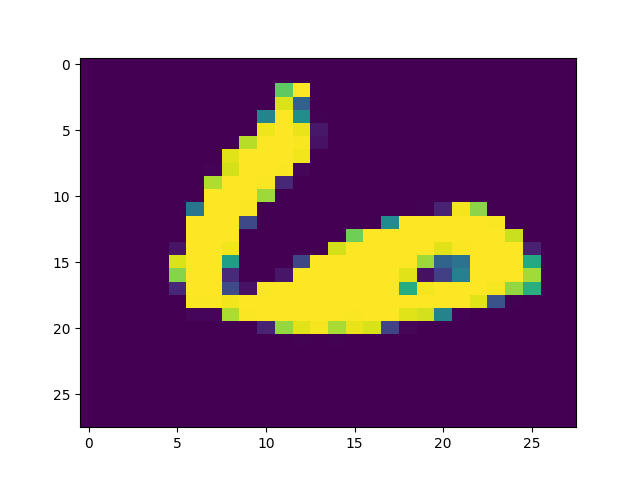}
    \caption{DAE \#6}
    \end{subfigure}%
     \begin{subfigure}{0.2\textwidth}
    \centering \includegraphics[scale=0.15]{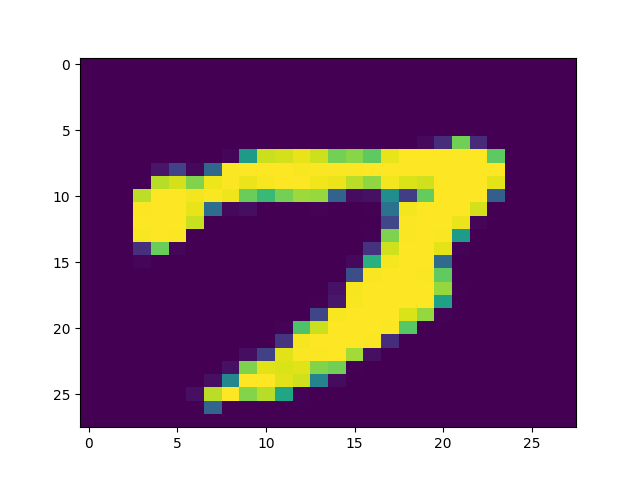}
    \caption{DAE \#7}
    \end{subfigure}%
     \begin{subfigure}{0.2\textwidth}
   \centering \includegraphics[scale=0.15]{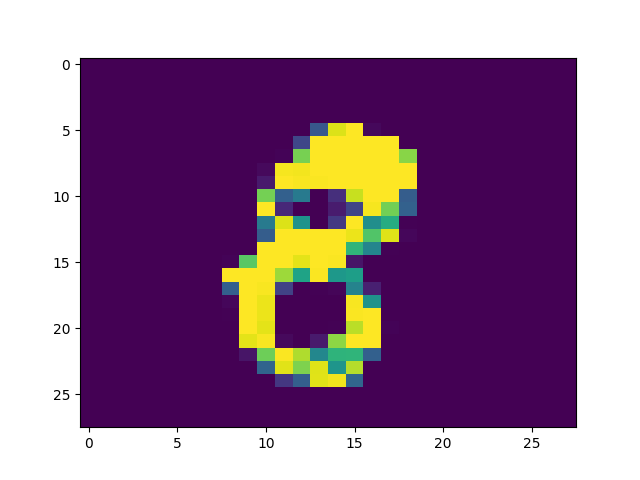}
    \caption{DAE \#8}
    \end{subfigure}%
    \begin{subfigure}{0.2\textwidth}
    \centering \includegraphics[scale=0.15]{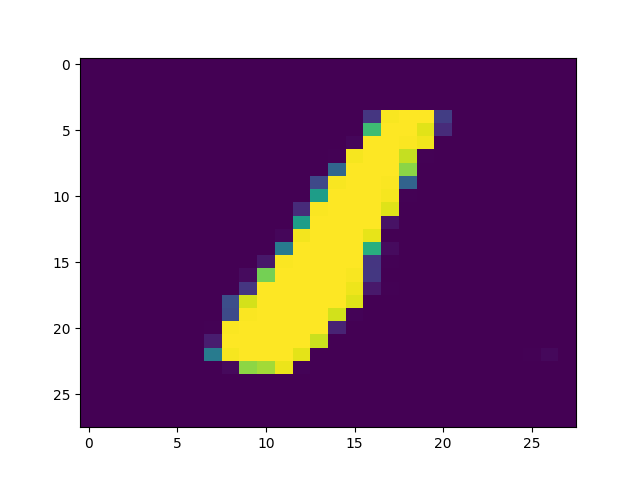}
    \caption{DAE \#9}
    \end{subfigure}%
    \caption{The outputs of the different DAEs with a vector of all-ones input.}
    \label{fig:synthetic_allones}
\end{figure*}

\begin{figure*}[t]
    \centering
    \begin{subfigure}{0.2\textwidth}
    \centering \includegraphics[scale=0.15]{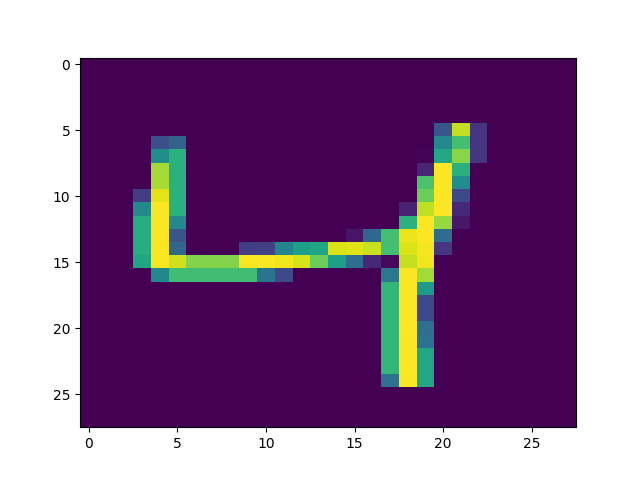}
    \caption{Input}
    \label{fig:4}
    \end{subfigure}%

     \begin{subfigure}{0.2\textwidth}
    \centering \includegraphics[scale=0.15]{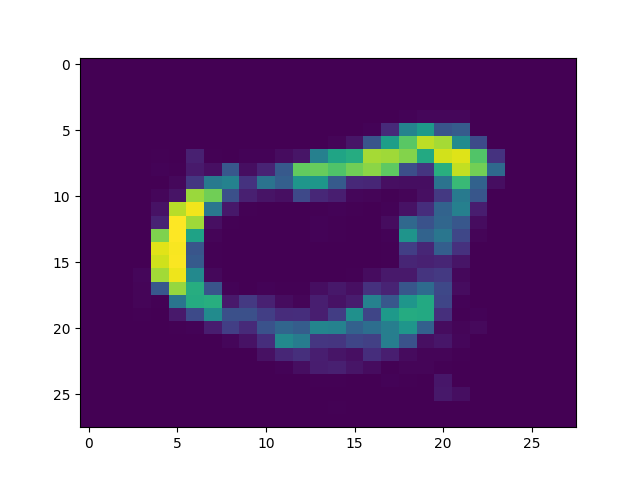}
    \caption{DAE \#0}
    \end{subfigure}%
     \begin{subfigure}{0.2\textwidth}
    \centering \includegraphics[scale=0.15]{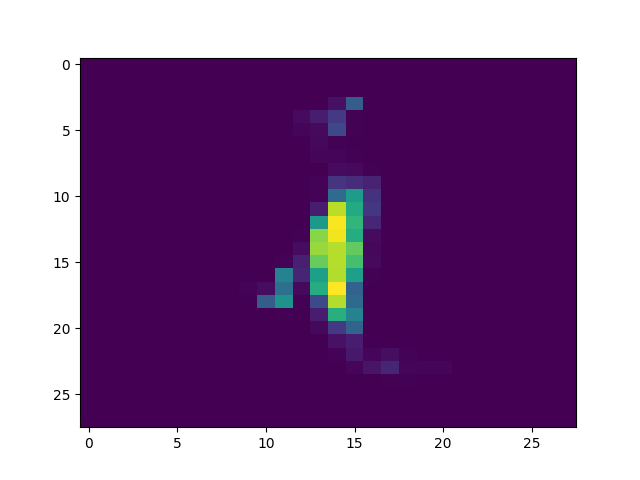}
    \caption{DAE \#1}
    \end{subfigure}%
     \begin{subfigure}{0.2\textwidth}
    \centering \includegraphics[scale=0.15]{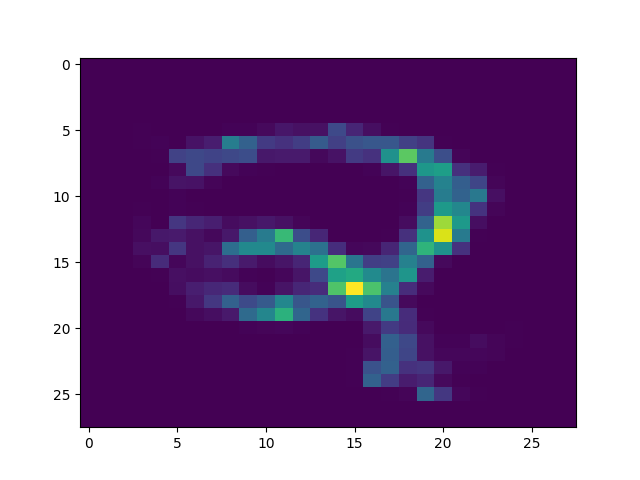}
    \caption{DAE \#2}
    \end{subfigure}%
    \begin{subfigure}{0.2\textwidth}
    \centering \includegraphics[scale=0.15]{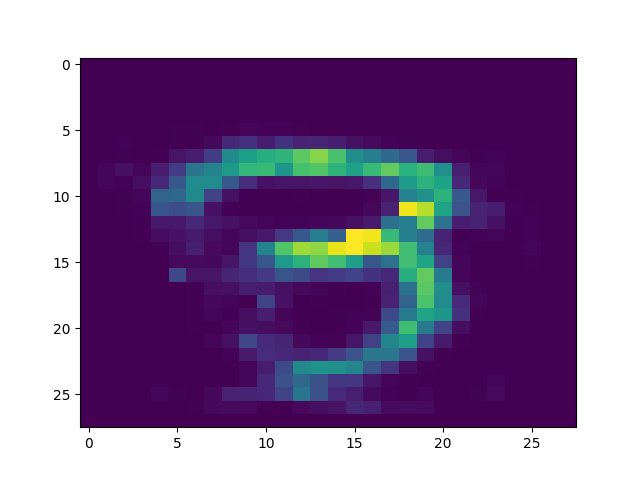}
    \caption{DAE \#3}
    \end{subfigure}%
     \begin{subfigure}{0.2\textwidth}
    \centering \includegraphics[scale=0.15]{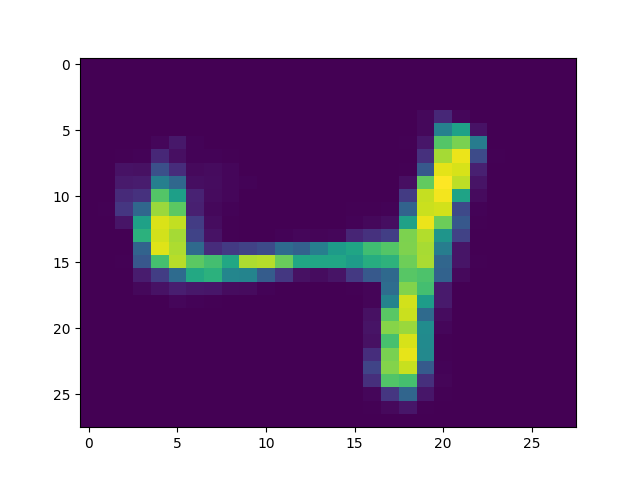}
    \caption{DAE \#4}
    \end{subfigure}%

    \begin{subfigure}{0.2\textwidth}
    \centering \includegraphics[scale=0.15]{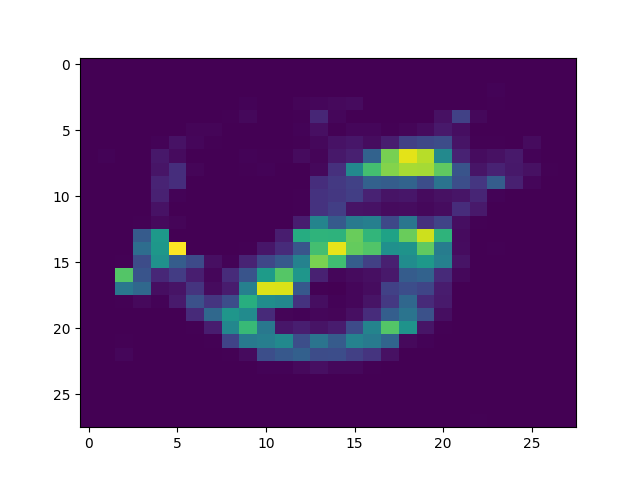}
    \caption{DAE \#5}
    \end{subfigure}%
     \begin{subfigure}{0.2\textwidth}
    \centering \includegraphics[scale=0.15]{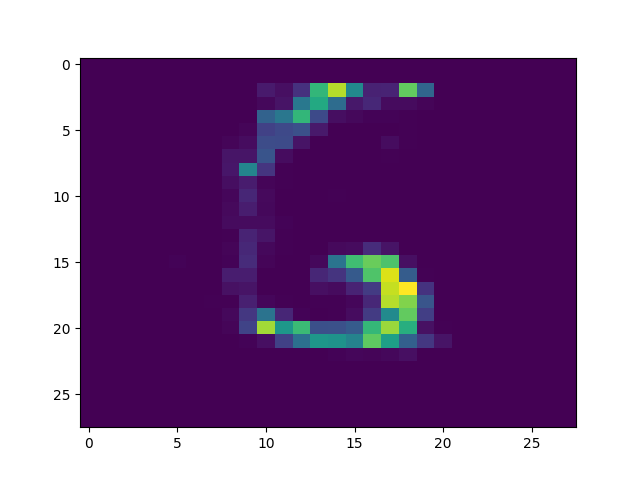}
    \caption{DAE \#6}
    \end{subfigure}%
     \begin{subfigure}{0.2\textwidth}
    \centering \includegraphics[scale=0.15]{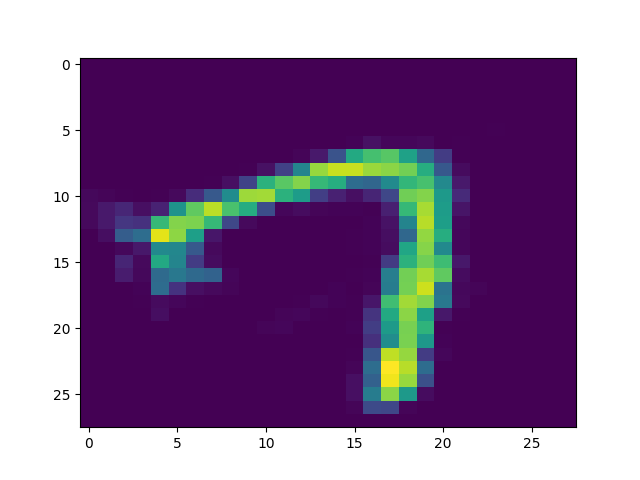}
    \caption{DAE \#7}
    \end{subfigure}%
     \begin{subfigure}{0.2\textwidth}
   \centering \includegraphics[scale=0.15]{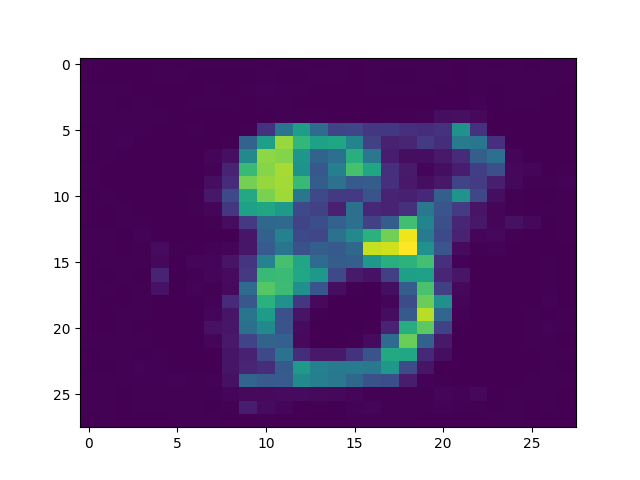}
    \caption{DAE \#8}
    \end{subfigure}%
    \begin{subfigure}{0.2\textwidth}
    \centering \includegraphics[scale=0.15]{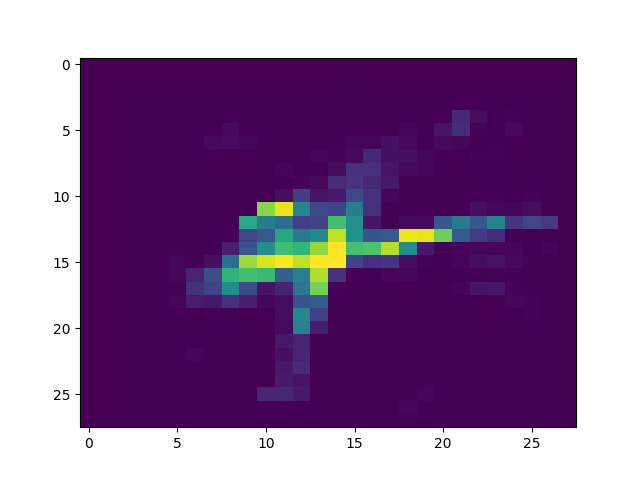}
    \caption{DAE \#9}
    \end{subfigure}%
    \caption{An example of the outputs of the different DAEs with the digit `4' as the input.}
    \label{fig:digit_4}
\end{figure*}
\subsection{MNIST}
The MNIST database has 70000 hand written gray-scale images of digits. Each image size is $28 \times 28$  pixels. Note that we worked on the raw data of the dataset (without pre-processing). For simplicity, the architecture of each one of the DAE was identical. Specifically, for the MNIST dataset we used a 5-layer network with   1024, 256, 10, 256, 1024 neurons,  respectively. The output layer of each DAE was set to be the sigmoid function.  For the clustering network we used simpler network with a 3-layer with  512, 512, 10 neurons, respectively.  The output transfer layer of the clustering network was the softmax function.
Table \ref{table:mnist_results} presents the results of the \ac{NMI}, the \ac{ARI} and  the \ac{ACC}  of the proposed \ac{DAMIC} method and several standard  baselines. It is clear that the \ac{DAMIC} outperforms the other methods on in all measures.  

\minisubsection{DAE expertise}
To test the expertise of each one of the DAE we conducted the following  experiment. After the clustering algorithm converged on the MNIST dataset, we synthetically created a new image in which all the pixels were set to be `1' (Fig.~\ref{fig:all_one}). The image reconstruction of all the 10  DAEs is shown in  Fig.~\ref{fig:synthetic_allones}.  It is evident that each DAE assumes  a different pattern of  input. Specifically, each DAE is responsible for a different digit. The clustering task was unsupervised and we sorted the autoencoders in Fig.  \ref{fig:all_one}) by their corresponding digits from `0' to `9' merely for purpose of visualization.

\minisubsection{Best reconstruction wins}
To further understand the behavior of the gate we carried out a different test. An image of the digit `4' was fed to the network (Fig.~\ref{fig:4}). The outputs of the different DAEs are  depicted in Fig.~\ref{fig:digit_4}.  Since each DAE specializes in a different digit, it was expected that the respective DAE would have  the lowest reconstruction error.  This was also reflected in  decision of the  clustering network $p(c=4|x; \theta_c)=0.99$. Note, that the other DAEs reshaped the reconstruction to be close to their digit expertise.

\begin{table*}
  \centering
    \caption{Ablation study on the MNIST database.}\label{table:ablation_results}
\begin{tabular}{l|*{5}c}
\hline
Method            &  \ac{DAMIC} & Pre-training only  & Joint-training only  & KM & \\
\hline
NMI              & \textbf{0.87} & 0.74 & 0.71  & 0.50   \\
ARI              & \textbf{0.81} & 0.67 & 0.53  & 0.37   \\
ACC              & \textbf{0.89} & 0.80 & 0.60  & 0.53   \\
\hline
\end{tabular}
\end{table*}

\subsection{Fashion}
The Fashion dataset shares the same dimensions and structure as the MNIST dataset. The only difference is the content of the images, which are now one of ten fashion items. Therefore, the same processing described in the former section was carried out on this dataset. 

Table~\ref{table:fashion_results} describes the results on the Fashion dataset. It is clear that the proposed DAMIC algorithm outperforms the compared algorithms. 

We note that the best reported clustering results
on the MNIST data achieved by the VaDE algorithm \cite{vade} that applies  variational autoencoder modeling assuming mixture of Gaussians distribution of the latent random variable. We compared our method to state-of-the-art deep  $k$-means-based algorithms using the same network architecture and parameter initialization and showed improvement in the performance. VaDE algorithm belongs to a different family of algorithms with different network architecture and parameter initialization strategies.
Hence, a direct performance comparison is difficult since it is heavily dependent on the implementations. It is worth noting that in the Fashion dataset, our results even outperforms the VaDE and the DEC with data augmentation (DEC-DA) algorithms results~\cite{decaugmentation2018} in this dataset.  

\subsection{20NEWS}
The 20Newsgroup corpus consists of 18,846 documents from 20 news groups. As in \cite{yang_2017} we also used the tf-idf representation of the documents and picked the 2,000 most frequently used words as the features. The architecture used in each one of the DAEs for this experiment also consisted of a 5-layer DNN with 1024, 256, 20, 256, 1024 neurons, respectively. The clustering network here consisted of 512, 512, 20 neurons.

Table \ref{table:20newsgroup_results} shows the results of the \ac{NMI}, \ac{ARI} and \ac{ACC} measures. It is clear that the proposed clustering method outperformed the competing baseline algorithms.

\subsection{RCV1}
The dataset used in this experiment is a subset of the RCV-1-v2 with 365, 968 documents, each containing one of 20 topics. As in \cite{yang_2017} the 2,000 most
frequently used words (in the tf-idf form) are used  as the features for
each documents. In contrast to the previous databases, in the RCV1 dataset, the size of each class is not equal. Therefore, KM-based approaches might not be  sufficient in this case.  In our architecture we used 1024, 256, 20, 256, 1024 \ac{elu} neurons in all DAEs, respectively, and in the clustering network we used 512, 512, 20 \ac{elu} neurons.

Table \ref{table:RCV1_results}  presents the 3 objective measurements for the RCV1 experiment. The proposed method outperformed the competing methods in NMI and ARI measures and had the same ACC score as of the DCN.

\nocite{Caron_2018} \nocite{Hsu_2018}

\subsection{Ablation study}
The DAMIC algorithm comprises of two steps, the initialization step, which is based on a deep autoencoder followed by a $k$-means clustering (DAE+KM), and a joint training of the gate and the clusters' autoencoders. We next explore the necessity of each part in the algorithm. For that, we compared the DAMIC with two variants of the proposed algorithm. The first one is based only on the initialization phase (without the joint the training), and the second is based only on the joint training phase with random initialization. 

Table~\ref{table:ablation_results} describes the results of our ablation study. The first conclusion is that the joint training with random initialization  improves the $k$-means results. This result confirm the first contribution of this paper that even with random initialization, the algorithm does not suffer from the clustering collapsing problem.  It is also evident that the pre-training phase is important for parameter initialization but it is not enough. The proposed algorithm which employs both, initialization step and joint training outperforms each of the variants separately.

\section{Conclusion}
In this study we presented a clustering technique which leverages the strength of deep neural network. Our technique has two major properties: first, unlike most previous methods, the clusters are represented by an autoencoder network instead of a single centroid vector in the embedded space. This enables a much richer representation of each cluster. Second, the algorithm does not cause a data collapsing problem. Hence, there is no need for regularization terms that have to be tuned for each dataset separately.
Experiments on a variety of real datasets showed the strong performance of the proposed algorithm over the other methods.


{\small
\bibliographystyle{ieee}
\bibliography{paper}
}

\end{document}